\documentclass[10pt,twocolumn,letterpaper]{article}

\PassOptionsToPackage{table,xcdraw}{xcolor}
\usepackage[pagenumbers]{cvpr} 
\usepackage{multirow}
\usepackage{algorithm}
\usepackage{algpseudocode}
\usepackage{verbatim}

\definecolor{cvprblue}{rgb}{0.21,0.49,0.74}
\usepackage[pagebackref,breaklinks,colorlinks,allcolors=cvprblue]{hyperref}

\newcommand{\CUT}[1]{}

\newcommand{\bz}{\boldsymbol{z}}

\def\paperID{17518} 
\def\confName{CVPR}
\def\confYear{2025}

\title{Advancing Multiple Instance Learning with Continual Learning \\ for Whole Slide Imaging}

\author{Xianrui Li\\
\small Dept. of Computer Science\\
\small City University of Hong Kong\\
{\tt\small xianrui.li@my.cityu.edu.hk}
\and
Yufei Cui\\
\small Noah's Ark Lab, Huawei Canada\\
\small Montreal, Quebec, Canada\\
{\tt\small yufei.cui@huawei.com}
\and
Jun Li\\
\small Guangzhou Bingli Technology Co., Ltd.\\
\small Guangzhou, Guangdong\\
\and
Antoni B. Chan\\
\small Dept. of Computer Science\\
\small City University of Hong Kong\\
{\tt\small abchan@cityu.edu.hk}
}

\begin{document}
\maketitle

\begin{abstract}
Advances in medical imaging and deep learning have propelled progress in whole slide image (WSI) analysis, with multiple instance learning (MIL) showing promise for efficient and accurate diagnostics. However, conventional MIL models often lack adaptability to evolving datasets, as they rely on static training that cannot incorporate new information without extensive retraining. 
Applying continual learning (CL) to MIL models is a possible solution, but often sees limited improvements. 
In this paper, we analyze CL in the context of attention MIL models and find that the model forgetting is mainly concentrated in the attention layers of the MIL model. 
Using the results of this analysis we propose two components for improving CL on MIL:
Attention Knowledge Distillation (AKD) and the Pseudo-Bag Memory Pool (PMP). AKD mitigates catastrophic forgetting by focusing on retaining attention layer knowledge between learning sessions, while PMP reduces the memory footprint by selectively storing only the most informative patches, or ``pseudo-bags'' from WSIs. Experimental evaluations demonstrate that our method significantly improves both accuracy and memory efficiency on diverse WSI datasets, outperforming current state-of-the-art CL methods. This work provides a foundation for CL in large-scale, weakly annotated clinical datasets, paving the way for more adaptable and resilient diagnostic models.
\end{abstract}    
\section{Introduction}
\label{sec:intro}
The advancement of AI technology has significantly transformed medical imaging, with notable progress in whole slide image (WSI) analysis \citep{yufei2023bayes}. In particular, developments in multiple instance learning (MIL) have enhanced the diagnostic accuracy and efficiency of deep learning models applied to WSIs \citep{lu2021data,zhang2022dtfdmil}, substantially reducing diagnostic time and minimizing misdiagnosis rates. Despite these advancements, most MIL methodologies currently rely on a static framework \citep{huang2023conslide,Esbri2024MICILMC}, wherein models are optimized using a fixed dataset. This static approach creates a critical limitation in terms of knowledge updating. As medical technology evolves rapidly, new data is continually generated \cite{8447230,10.1001/jama.2017.14585,litjens20181399}, often with variations in imaging equipment, acquisition methods, and standardized settings. These factors alter the labels and distribution patterns of new datasets, making it essential for models to adapt to these changes \citep{takahama2023domain}. Full retraining of models using both new and historical data, however, is resource-intensive and time-consuming. Consequently, continual learning (CL) \citep{wang2024comprehensive,hadsell2020embracing} emerges as an efficient solution to this challenge, enabling models to incrementally integrate new information and thereby maintain their diagnostic relevance without the prohibitive costs of full retraining.
However, CL faces catastrophic forgetting \citep{kirkpatrick2017overcoming}, where models lose previously acquired knowledge when learning new information, addressed by regularization-based, architectural-based, and replay-based methods \citep{van2019three}.

Our research focuses on class-incremental learning (CIL), a specialized area within CL that holds particular promise for applications in medical imaging. As medical knowledge advances, disease annotations are becoming increasingly detailed, leading to a proliferation of diagnostic categories. This development creates a pressing need for models capable of continually incorporating new classes of medical information \citep{Esbri2024MICILMC} while retaining previously acquired knowledge, i.e., 
class-incremental learning. 
Moreover, research suggests that models trained across multiple medical domains often achieve superior performance compared to those restricted to single domains \citep{gao2024accurate}, underscoring the potential of multi-domain learning to enhance diagnostic accuracy and generalizability.

Applying CL 
to WSI presents unique challenges due to the substantial data complexity and annotation demands associated with WSIs. 
Each WSI contains gigabytes of pixel data, making detailed annotations costly and time-consuming due to required expert input. Most WSIs only have slide-level labels without detailed annotations. To address this limitation, MIL has become a widely adopted strategy in WSI analysis. In the MIL paradigm, each WSI is conceptualized as a ``bag'' of smaller ``patches'' allowing models to be trained using high-level, whole-slide labels rather than exhaustive pixel- or patch-level annotations.
In particular, \emph{attention} MIL \citep{ilse2018attentionbased,lu2021data,shao2021transmil,zhang2023attentionchallenging} first uses an attention network to obtain an attention weight on each patch, with which a slide-level feature vector is calculated via the weight and the patch-level features. Finally, the slide-level feature serves as the input to a slide-level classifier.

Recent advancements 
have begun to explore the integration CIL within the MIL framework, with encouraging results \citep{huang2023conslide, Esbri2024MICILMC}. Some studies have proposed new frameworks specifically designed for CIL in WSI, yet these often require substantially more parameters and show only moderate performance improvements in conventional settings. Other approaches attempt to adapt existing CIL techniques for use with MIL models, also yielding only limited enhancements. Through our investigation of MIL under CIL, we have identified a critical insight: the patterns of forgetting in attention MIL classification models under CIL differ significantly from those observed in traditional image classification models. Specifically, our theoretical analysis and model visualizations indicate that \emph{model forgetting in MIL is concentrated primarily in the attention network}, rather than in other layers like the classifier (logit) layer.

Based on our findings, we propose two effective and adaptable methods that improve the performance of traditional MIL models in 
CIL scenarios. The first method, attention knowledge distillation (AKD), addresses forgetting by imposing constraints on the attention layers during CL training. By using distillation techniques to transfer knowledge from previous models to the current one, AKD balances the retention of prior knowledge with the assimilation of new information. The second method, the pseudo-bag memory pool (PMP), mitigates the high memory demands of replaying WSI data by using a patch importance scoring technique to selectively sample critical patches, thereby substantially reducing memory requirements. Together, AKD and PMP provide a robust framework for enhancing MIL in CL settings, addressing both knowledge retention and memory efficiency while accommodating the specific challenges posed by WSIs.

In summary, our work presents three major contributions to MIL and continual learning research:

\begin{itemize}
    \item To the best of our knowledge, this is the first attempt to theoretically and experimentally explore the fundamental reasons for the catastrophic forgetting of attention MIL in CIL. Compared to classical classification models where forgetting occurs mainly in the classification layer, we show that attention MIL experiences forgetting primarily in the attention layer.
    \item We have developed a specialized knowledge distillation strategy tailored for attention networks in MIL, which significantly enhances model performance during CL, empirically approaching the upper limit achievable through joint training.
    \item We introduce a novel pseudo-bag memory pool method, which effectively facilitates the implementation of replay-based approaches in MIL, offering a new direction for memory management in this context.
\end{itemize}

\section{Related Work}
\label{sec:related}

\subsection{Multiple Instance Learning (MIL)}
Many MIL models leverage bag representations to predict the slide label. For instance, ABMIL \citep{ilse2018attentionbased} generates a bag representation by calculating
weighted patch features,
enabling it to focus on the most informative patches within a slide. Building on this approach, CLAM \citep{lu2021data} 
refines this with instance clustering for multi-category tasks.
TransMIL \citep{shao2021transmil} proposes a correlated MIL framework that integrates multi-head self-attention and spatial information to capture global correlations across the an image. Similarly, DTFD \citep{zhang2022dtfdmil} employs a two-tier framework with ABMIL as the foundational module in each tier, aiming to capture hierarchical features across different levels of granularity. 

While these models demonstrate strong performance on static datasets, they generally do not account for scenarios involving non-stationary data. In dynamic environments where data distributions may shift over time, traditional MIL approaches such as CLAM \citep{lu2021data} and TransMIL \citep{shao2021transmil} are vulnerable to issues like catastrophic forgetting.

\subsection{Continual Learning (CL)}
CL methodologies in computer vision 
are generally categorized into three main approaches. The first, regularization-based methods, seek to address catastrophic forgetting by constraining the gradients of model parameters like EWC \citep{kirkpatrick2017overcoming} and MAS \cite{aljundi2017memory}. However, these methods often produce suboptimal results in complex environments like realistic pathology data. The second category, model architecture-based methods, involves modifying network structures through strategies like compression and expansion to incorporate new knowledge without overwriting existing information. Approaches such as PNN \citep{rusu2016progressive} and PackNet \citep{mallya2017packnet} exemplify this technique, which, while theoretically robust for retaining prior knowledge, tends to suffer from scalability issues as an extension of the MIL model is not as simple as the classical classification model. Finally, replay-based methods utilize selective reintroduction of past data samples to reinforce memory, helping models to maintain prior knowledge during the training of new tasks, like DER \citep{buzzega2020dark} and GEM \citep{lopez2017gradient}. Nevertheless, these methods face challenges related to memory pool configuration and sample selection, which are critical since WSI datasets only contain hundreds of samples, and a WSI may be too large to store in the memory.

\subsection{Class-incremental learning in MIL} \label{sec:aug}
Recent studies have increasingly explored the application of class-incremental learning (CIL) within MIL.
ConSlide \citep{huang2023conslide} 
combines fast and slow learning, substituting the WSI buffer with a region buffer. This adaptation effectively reduces memory pool requirements and enables efficient data replay. Alternatively, MICIL \citep{Esbri2024MICILMC} adopts a regularization-based strategy that builds on the LwF \citep{li2016learning} paradigm.

In contrast to these approaches, through theoretical analysis and model visualizations, we identify that model forgetting is primarily concentrated in the \emph{attention layers}. Based on our findings, we then propose two new methods that focus on preserving knowledge learned by the attention network, which provide a significant boost to MIL performance in CIL scenarios.


\section{Motivation}
\subsection{Problem Statement}
\textbf{MIL.} In the WSI MIL classification problem, the WSI ${\cal X}$ comprises a bag of patches, $\mathbf{P} = \{\mathbf{p}_n\}^N_{n=1}$, and is associated with the bag-label $y$, where $N$ is the number of patches, which varies with different bags. The objective of MIL is to 
predict
the bag-level $y$ label with the patches in the bag.

We follow the standard formulation of ABMIL \citep{ilse2018attentionbased}. ABMIL classifies a WSI in three steps following the MIL paradigm. First, a WSI is divided into $N$ patches, and a bag of patch features is obtained via a fixed feature extractor $g(\cdot)$, $\mathbf{H} = \{\mathbf{h}_n\}^N_{n=1}$, where the $n$-th patch feature is $\mathbf{h}_n = g(\mathbf{p}_n) \in \mathbb{R}^d$, and $d$ is the dimension of the feature vector. Second, the patch features are aggregated to compute a bag-level (slide-level) feature $z$:

\begin{equation}
  \mathbf{z} = \mathbf{H} \mathbf{a} = \sum\nolimits_{n=1}^N a_n \mathbf{h}_n, 
  \label{eq:1}
\end{equation}
where $a_n$ is the normalized attention score of the $n$th patch,
\begin{equation}
  a_n = \frac{\exp\left\{ \mathbf{w}^\top (\tanh(\mathbf{V}_1 \mathbf{h}_n) \circ \sigma(\mathbf{V}_2 \mathbf{h}_n)) \right\}}{\sum_{m=1}^N \exp\left\{ \mathbf{w}^\top (\tanh(\mathbf{V}_1 \mathbf{h}_m) \circ \sigma(\mathbf{V}_2 \mathbf{h}_m)) \right\}},
  \label{eq:attn}
\end{equation}
and $\theta=(\mathbf{w}, \mathbf{V}_1, \mathbf{V}_2)$ are learnable weight matrices, $\tanh$ and $\sigma$ refer to the hyperbolic tangent and sigmoid activation functions, and $\circ$ denotes the element-wise product. We denote $\mathbf{a} = [a_n]_n = f_\theta(\mathbf{H})$ as the attention vector, where $f_\theta$ is the attention network.
Third, the bag label is predicted by a classifier $\hat{y} = f_\phi(\mathbf{z})$, where $f_\phi$ is an MLP.

\noindent\textbf{CIL.} 
CIL is a  CL 
scenario consisting of training a model on tasks with disjoint label spaces. We define a sequence of datasets ${\cal D} = \{{\cal D}_1, {\cal D}_2, \cdots, {\cal D}_T\}$, where the $t$-th dataset ${\cal D}_t = \{{\cal X}, y, t\}_k$ contains tuples of the WSI sample ${\cal X}$, its corresponding label $y \in {\cal Y}_t$ (where ${\cal Y}_i \cap {\cal Y}_j = \emptyset$ for $i \neq j$), and the task identifier $t$ (available only during training). The subscript $k$ means there are $k$ samples for dataset ${\cal D}_t$.
In CIL, 
the model needs to predict the classification label $y$ with WSI ${\cal X}$ as the input.

\begin{table}[]
\centering
\caption{Test Accuracy of CLAM and TransMIL on Task 1 using different combinations of its attention/classifier from CL task sessions ($t$). The diagonal cells (blue) correspond to the accuracy of fine-tuned model $(\theta_t,\phi_t)$ after each task session.}
\label{tab:decouple-clam}
\small
\begin{tabular}{ll|lll}
\multicolumn{2}{c|}{} & \multicolumn{3}{c}{Attention $\theta_t$}                                     \\
\multicolumn{2}{c|}{\multirow{-2}{*}{CLAM}} &
  \multicolumn{1}{c}{$t$=1} &
  \multicolumn{1}{c}{$t$=2} &
  \multicolumn{1}{c}{$t$=3} \\ \hline
         & $t$=1        & \cellcolor[HTML]{CBCEFB}0.8621 & 0.1552                         & 0.0517 \\
         & $t$=2        & 0.8793                         & \cellcolor[HTML]{CBCEFB}0.0000 & 0.0000 \\
\multirow{-3}{*}{\begin{tabular}[c]{@{}l@{}}Classifier\\ $ \phi_t$\end{tabular}} &
  $t$=3 &
  0.8448 &
  0.0000 &
  \cellcolor[HTML]{CBCEFB}0.0000 \\ \hline
\end{tabular}

\vspace{0.2cm}
\begin{tabular}{ll|ccc}
\multicolumn{2}{c|}{}                           & \multicolumn{3}{c}{Attention $\theta_t$}                                     \\
\multicolumn{2}{c|}{\multirow{-2}{*}{TransMIL}} & $t$=1                            & $t$=2                            & $t$=3    \\ \hline
                      & $t$=1                     & \cellcolor[HTML]{CBCEFB}0.8966 & 0.0345                         & 0.0345 \\
                      & $t$=2                     & 0.8966                         & \cellcolor[HTML]{CBCEFB}0.0345 & 0.0000 \\
\multirow{-3}{*}{\begin{tabular}[c]{@{}l@{}}Classifier\\ $ \phi_t$\end{tabular}} & $t$=3 & 0.8276 & 0.0345 & \cellcolor[HTML]{CBCEFB}0.0000 \\ \hline
\end{tabular}
\end{table}

\subsection{Decoupling Experiments}
\label{sec:decouple}
In this section, we investigate the cause of the performance degradation of CIL for MIL. Here we consider CLAM \citep{lu2021data} and TransMIL \citep{shao2021transmil} as a representative attention  MIL. We define the attention network $\theta$ as the sub-network that computes the attention scores from the patch features \cref{eq:attn} for CLAM and \citet{xiong2021nystrmformer} for TransMIL), and we define the classifier  $\phi$ as the sub-network that takes the bag-level features $\mathbf{z}$ and predicts the class label $y$.

In our CL experiment after the $T$-th task session, we obtain a set of $T$ fine-tuned model $\{(\theta_t, \phi_t)\}_{t=1}^T$, which represents the evolution of the model when sequentially fine-tuning with new task data. 
To investigate the effect of CL, we then recombined the trained attention network $\theta_t$ and classifier $\phi_{t'}$ from different CL task sessions, and tested these ``hybrid" models on Task 1 test data.

\begin{figure}
    \centering
    \includegraphics[width=1.\linewidth]{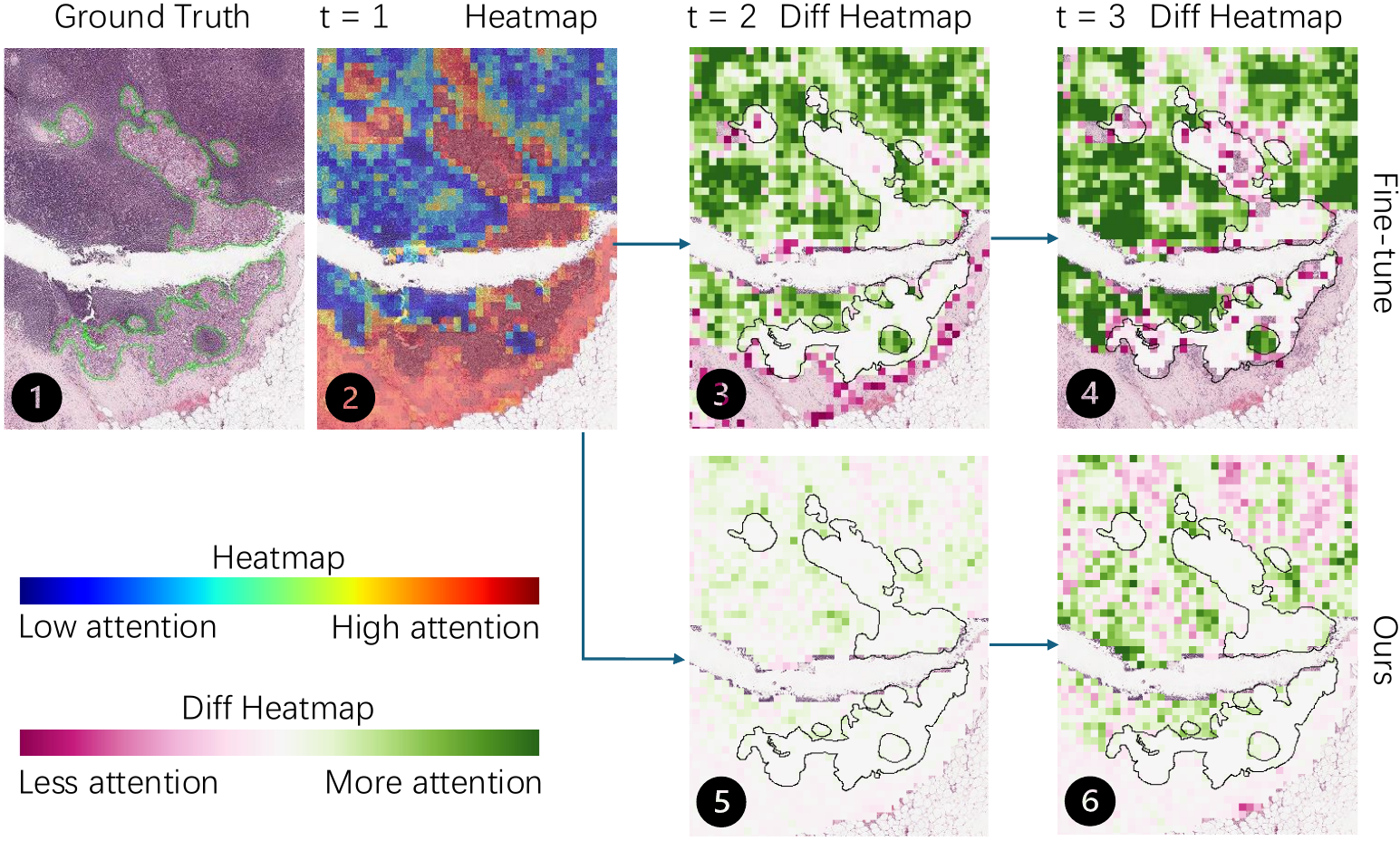}
    \caption{Example of the attention shift during CL for a Task 1 test image. (1) Input image with tumor area outlined in green. (2) The attention heat map for $t$=1 session. (3) and (4): the attention drift, relative to (2), after CL sessions $t \in \{2, 3\}$ for fine-tuning CL. The tumor areas are outlined in black. 
    (5) and (6): the attention drift for our proposed method. Our method better preserves the attention much better than fine-tuning.}
    \label{fig:vis}
\end{figure}

We use Camelyon-TCGA for the benchmark, as described in the \cref{sec:exp-setup}. The results for all pairs of attention/classifier are presented in
\cref{tab:decouple-clam}.
The significant decline of Task 1 accuracy after training on subsequent tasks is evident in the diagonal cells of the tables. 
However, examining the 1st columns of the tables, we find that the performance does not decline much if the original attention layer $\theta_1$ is used to replace the attention layers learned after subsequent tasks (i.e., used with classifier $\phi_t$).
Furthermore, if the classifier after each task is replaced with the original $\phi_1$, there is still a significant decline in Task 1 accuracy (in the 1st row of the tables).
These two observations suggest that \emph{the drift of the fine-tuned attention network is the main cause of the performance decline of attention MIL in CL}. \cref{fig:vis} (2-4) illustrate the drift of CLAM's attention during subsequent CL sessions, where the attention on the tumor region (outlined region) is reduced after each CL session.

\subsection{Analysis of Parameter Update in MIL} \label{sec:theory}
To explore the underlying mathematical mechanisms of catastrophic forgetting, we 
examine the gradient shifts in the attention network and classifier in the MIL model during training updates. As a foundational example, we employed the ABMIL \cite{ilse2018attentionbased} model,
where
the forward propagation is:
\begin{equation}
f_\phi = \boldsymbol{\phi}^\top \mathbf{Ha},
\end{equation}
where \(\boldsymbol{\phi} = [\phi_1, \cdots, \phi_d]^\top\) represents the logit layer of the binary classifier, \(\mathbf{a} = [a_1, \cdots, a_n]^\top\) are the attention scores from \cref{eq:attn}, and \(\mathbf{H} = [\mathbf{h_1}, \cdots, \mathbf{h_n}]\) are the feature vectors corresponding to patches in the WSI.
In the binary classification setting, the model's loss function is
\begin{equation}
L = \log(1 + e^{-y f_\phi}).
\end{equation}

From this, the gradients w.r.t.~each parameter in the model can be derived (see supplemental). 
The squared norm of the gradient w.r.t.~each classifier weight \(\phi_j\) is: 
\begin{equation}
\big(\tfrac{\partial L}{\partial \phi_j}\big)^2 = y^2 \sigma^2(-y f_\phi) z_j^2
\label{eq:grad-phi}
\end{equation}
where \( z_j = \sum_{i=1}^n a_i H_{ij} \) represents the \( j \)-th element of the aggregated feature vector \( \mathbf{z} \) and $H_{ij}$ is the $ij$-th element of $\mathbf{H}$.
Similarly, the squared norm of the gradient w.r.t.~each attention score \(a_i\) is:
\begin{equation}
\big(\tfrac{\partial L}{\partial a_i}\big)^2 = y^2 \sigma^2(-y f_\phi) (\boldsymbol{\phi}^\top \mathbf{h}_i)^2.
\label{eq:grad-attn}
\end{equation}
Through \cref{eq:grad-phi} and (\ref{eq:grad-attn}), we analyze the magnitude of the gradients with respect to different parameters in the attention network and the classifier. 
The main difference between the two equations are the terms $z_j^2$ and $\boldsymbol{\phi}^\top \mathbf{h}_i$.
For the gradient of the classifier in \cref{eq:grad-phi}, 
since the normalized attention $\mathbf{a}$ sums to 1 and $\mathbf{H}$ is fixed for an image, then $z_j$ is upper-bounded, $\max z_j = \max_i H_{ij}$.
Thus each update step of $\phi_j$ is always constrained within a specific range, 
which leads to the the classifier's parameters only gradually updating during training. In the context of CL, when there is a significant shift in the data stream, like tasks switch, the classifier only undergoes slow updates, thereby retaining more of the previous knowledge.

In contrast, the attention gradient in (\cref{eq:grad-attn})
directly reflects the importance 
the patch $\mathbf{h}_i$ to the classifier $\boldsymbol{\phi}$, and this gradient is unbounded -- as the classifier becomes more confident ($\|\boldsymbol{\phi}\|^2$ increases), then the attention gradients increase. Therefore, 
when there is a significant change in the data stream during CL, i.e., $\mathbf{h}_i$ undergoes drastic changes, the gradient of the attention layer will also fluctuate significantly, especially on newly-seen patches with high logit scores ($\boldsymbol{\phi}^\top \mathbf{h}_i$), leading to catastrophic forgetting due to rapid model updates.
Furthermore, if the newly seen patch with a high logit score is not informative or noisy, then the resulting large attention gradient makes the attention network overfit to this patch, causing generalization problems. 

\begin{figure}
    \centering
    \includegraphics[width=1\linewidth]{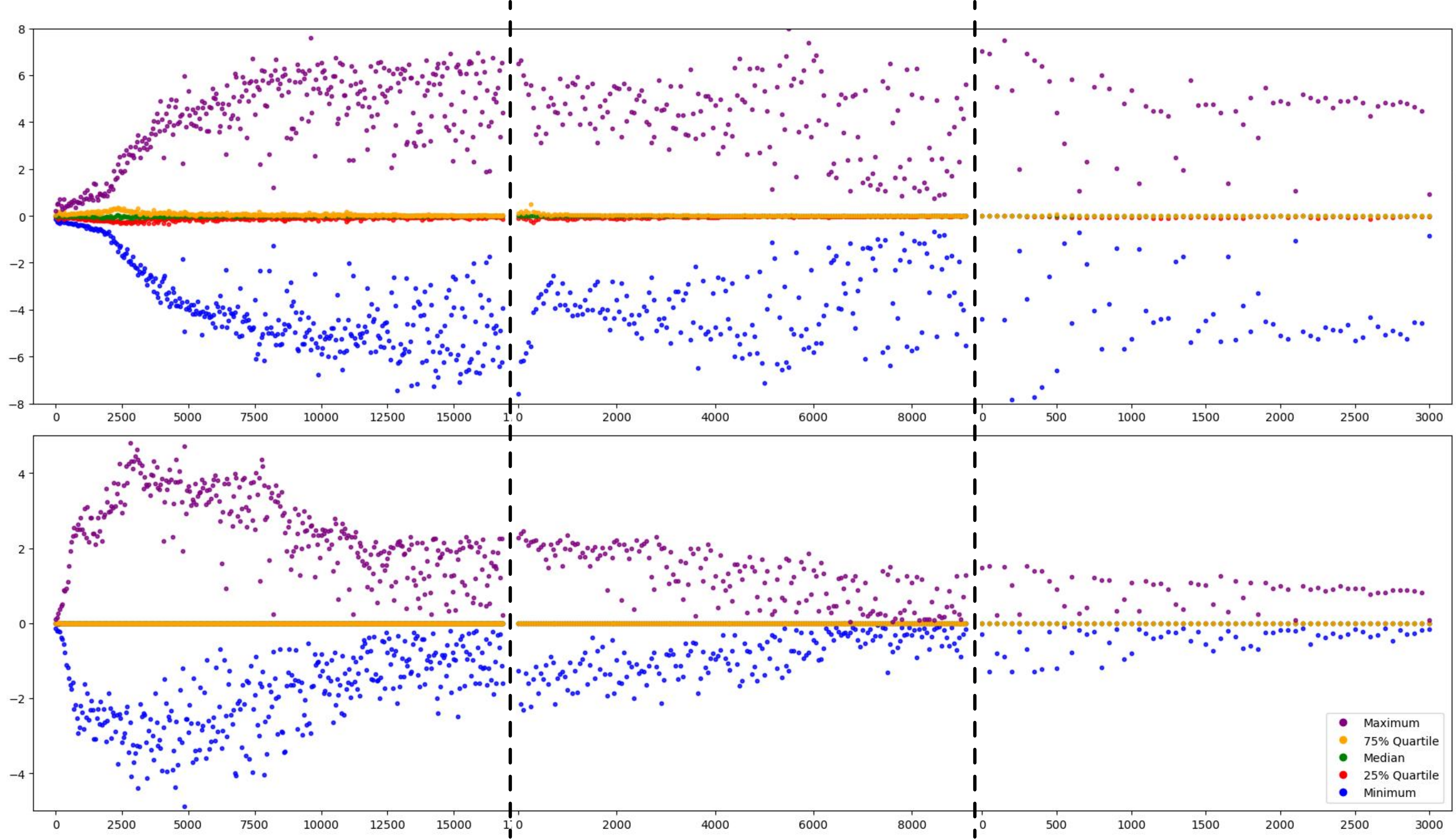}
    \caption{Evolution of the distribution of 
    gradient values of CLAM during CL for: 
(top) the attention network, (bottom) the classifier network.
The data points are aggregated where each dot represents the mean of 50 consecutive training steps. The five colored dots at each aggregated step represent the minimum, maximum, and three quartiles (first, second, and third) of the gradient values. The vertical black dotted lines mark the transitions between tasks.
}
    \label{fig:step-grad}
\end{figure}

In summary, the gradients of the classifier are bounded, while those in the attention layer (and the preceding attention network) are not. Thus, the attention network will be updated faster than the classifier, i.e., the loss decreases more when updating the attention network. \cref{fig:step-grad} shows the distribution of gradient values of the attention network and the classifier of CLAM as it evolves over successive training steps in CL. 
%
During training 
on the second and third tasks, the maximum and minimum values of the gradient of the classifier gradually shrink and stabilize, but the gradient of the attention network keeps oscillating in a higher range.
%
In the traditional image classification, \citet{DBLP:journals/entropy/KesslerCRZR23} found that the performance of CL
largely depends on different classification heads across tasks. However, for  MIL, we find that switching classification heads will no longer be effective, and our decoupling experiments in \cref{sec:decouple} further validate this theoretical finding.

\section{Methodology} \label{sec:methods}

In this section, using our analysis in the previous section, we propose two methods for CL of attention-based MIL.

\begin{figure*}
    \centering
    \includegraphics[width=.8\linewidth]{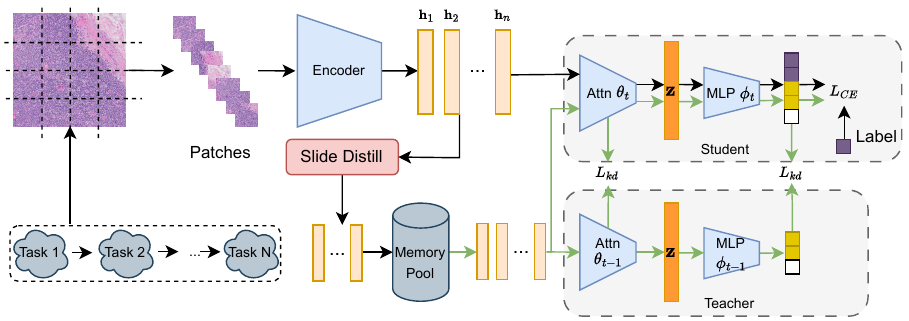}
    \caption{Overview of our framework architecture. The black arrows denote the forward pass for current data, including patch-wise slide processing, feature encoding, and memory pool updates. Orange rectangles represent patch-level features, while deep orange rectangles indicate bag-level feature representations. The green paths indicate the knowledge distillation process between teacher and student networks, incorporating both attention-based and logit-based distillation losses alongside cross-entropy supervision.}
    \label{fig:overview}
\end{figure*} 

\subsection{Attention Knowledge Distillation}
Inspired by Sec.~\ref{sec:theory}, we propose a 
 plug-and-play method called \emph{attention knowledge distillation} (AKD). Compared to the commonly used logits distillation in current CL approaches \citep{buzzega2020dark,li2016learning}, we
 focus on distillation of the attention layers since they drift faster in the MIL setting.
  Moreover, AKD can be combined with logits distillation to further enhance model performance.
Previous CL works used the output logit of the classifier $f_\phi(P)$ from previous tasks to mimic the original response for old samples. Therefore, their objective function for distillation is:
\begin{equation}
\mathcal{L}_{logits} = \mathbb{E}_{\mathbf{z} \sim \mathcal{D}_{t-1}} \left[ D_{KL}(f_{\phi_{t-1}}(\mathbf{z}) \parallel f_{\phi_t}(\mathbf{z})) \right],
\end{equation}
where $\phi_{t-1}$ are the optimal parameters at the end of task $t-1$,  $\phi_t$ are the current parameters being optimized, and $D_{KL}$ is the KL divergence. 

AKD directly optimizes the attention network using attention scores instead of the output logits of the model, reducing the interference caused to the classification layer. Specifically, consider bag feature $\mathbf{z} = \sum_n a_n \mathbf{h}_n$, AKD aims to ensure that after the model learns a new task, it retains the relative order of patch attention scores when inferring on previous tasks. In other words, after completing training on a new task, patches from previous tasks with previously high attention scores should continue to have high attention scores, and those with low scores should maintain relatively low scores. Therefore, we use KL divergence to enforce this relationship of the attention scores:
\begin{equation}
\mathcal{L}_{attn} = \mathbb{E}_{\mathbf{H} \sim \mathcal{M}} \left[ D_{KL}(f_{\theta_{t-1}}(\mathbf{H}) \parallel f_{\theta_t}(\mathbf{H})) \right],
\label{eq:kl-loss}
\end{equation}
where $\theta_{t-1}$ is the optimal attention network at the CL session $t-1$, and $\mathbf{H}$ are patch features sampled from the memory pool $\mathcal{M}$. Note that \( {\cal A} = f_{\theta_t}(\mathbf{H}) \) represents the output produced by the attention network, which may encompass more than just the raw attention scores. Depending on the specific architecture of the network, the output can include various feature vectors derived from the attention process. For instance, ${\cal A} = \{QK, QKV\}$ in the case of self-attention in TransMIL.

\subsection{Pseudo-Bag Memory Pool} \label{sec:meth-pmp}

Traditional replay-based methods typically store thousands of samples in the memory pool -- each sample occupies very little space, usually sized at 224x224 pixels, 
and having a large number of samples ensures that during replay training the model does not overfit.
However, in WSI 
tasks, the entire dataset contains only a few hundred images due to the high costs of data collection and patient privacy concerns. Moreover, each WSI encompasses tens of billions of pixels, making it impractical to store too many raw WSIs directly in the memory pool.

Recent research indicates that attention MIL models primarily utilize a small subset of patches from a WSI when calculating the bag feature $\mathbf{z}$ \citep{yan2023shapley,zhang2023attentionchallenging}. This suggests that the original slide can be distilled by removing low-contribution patches: $\mathbf{P} = \{\mathbf{p}_n\}_{n=1}^N \rightarrow \hat{\mathbf{P}} = \{\mathbf{p}_n\}_{n=1}^K$, where $K \ll N$. \citet{yan2023shapley} demonstrated that the top few dozen patches with the highest attention scores account for over 90\% of the model's attention. Therefore, one possible approach is to use patches with high attention scores (top-$K$) to approximate the original bag feature:

\begin{equation}
\hat{\mathbf{z}} = \sum\nolimits_{n=1}^K a_n \mathbf{h}_n \approx \mathbf{z} = \sum\nolimits_{n=1}^N a_n \mathbf{h}_n.
\end{equation}

Thus, instead of storing the entire slide, we store the distilled slide (pseudo-bag) in the memory pool $\mathcal{M} = \{(\hat{\mathbf{P}}_i, y_i)\}$ to approximate the data distribution of the WSI memory pool $\mathcal{M} = \{(\mathbf{P}_i, y_i)\}$. While using only the top-K patches is a tempting approach, it has been shown that other patches can also contribute valuable information \citep{zhang2022dtfdmil}.

Therefore, we propose the MaxMinRand strategy to distill the essential information from the WSI:

\begin{itemize}
    \item \textbf{MaxMinRand}: selects $K/2$ patches randomly and selects $K/4$ patches with the highest and the lowest attention scores, respectively.
\end{itemize}

We also consider other methods and will present them in \cref{sec:exp-abl}. We adopt the reservoir sampling method \citep{Vitter1985RandomSW} to update the slide-level memory pool. The reservoir sampling method can guarantee each slide-level sample shares the same probability of being stored and removed in the memory pool for the current task.

\subsection{Framework Overview}
The model's overall architecture, as depicted in Figure \ref{fig:overview}, follows a structured, task-oriented data flow. Input data is sequentially fed into the model in alignment with the progression of tasks. At the conclusion of each task, a subset of distilled samples is retained in a buffer through the PMP (Pseudo-Bag Memory Pool) method. When training transitions to the next task, the model is provided with data from both the current task and the buffered samples. For data from the current task, the model engages in regular parameter updates, as our experiments suggest that interference from prior tasks can impede effective adaptation to new data. During the replay of buffered samples from previous tasks, the model applies AKD to mitigate forgetting of prior knowledge. The final loss function, integrating these components, is formulated as follows:

\begin{equation}
\mathcal{L} = \mathcal{L}_{CE} + \alpha \mathcal{L}_{attn} + \beta \mathcal{L}_{logits},
\end{equation}
where $(\alpha,\beta)$ are hyperparameters that apply only to the data in the buffer. The complete process is outlined in Alg. \ref{alg:mil-cl}.

\begin{algorithm}[t]
\scriptsize
\caption{Overview}
\label{alg:mil-cl}
\begin{algorithmic}[1]
\State \textbf{Input:} Attention network \(f_{\theta}\), classifier \( f_{\phi}\), number of tasks \( T \), training set \( \mathcal{D} = \{\{(\mathbf{H}_i^t, y_i^t)\}_{i=1}^{n_t}\}_{t=1}^T \), \( y_i^t \in \mathcal{Y} \), memory pool $\mathcal{M}$.
\State \textbf{Initialize:} \( f_{\theta} \), \( f_{\phi} \), \( \mathcal{M} = \emptyset \)
\For{$t = 1$ to $T$}
    \State Obtain current task data $\mathcal{D}_t=\{(\mathbf{H}_i^t, y_i^t)\}_{i=1}^{n_t}$.
    \State Combine current task and memory pool \( \bar{\mathcal{D}} = \mathcal{D}_t \cup \mathcal{M} \).
    \For{ sample in \( \bar{\mathcal{D}} \)}
        \State Obtain $(\mathbf{H}, y)$ from sample
        \State Calculate attention outputs:  ${\cal A}, \bz=f_{\theta}(\mathbf{H})$.
        \State \(\mathcal{L}_{CE} = \mathcal{L}_{CE}(f_{\phi}(\bz), Y) \).
        \If{$(\mathbf{H}, y) \in \mathcal{M}$}
            \State Obtain previous output $(\tilde{\cal A},\tilde{f}_{\phi}(\bz))$ from sample.
            \State \(\mathcal{L}_{attn} = \mathcal{L}_{KL}({\cal A}, \tilde{\cal A}) \) (Eq.~\ref{eq:kl-loss}). 
            \State $\mathcal{L}_{logits} = \mathcal{L}_{KL}(f_{\phi}(\bz),\tilde{f}_{\phi}(\bz))$
        \Else
            \State $\mathcal{L}_{attn}=0, \mathcal{L}_{logits}=0$.
        \EndIf
        \State $\mathcal{L} = \mathcal{L}_{CE} + \alpha \mathcal{L}_{attn} + \beta \mathcal{L}_{logits}$.
        \State $(\theta, \phi) = (\theta, \phi) + \eta \nabla_{\theta, \phi}(\mathcal{L})$
        \State $(\tilde{\mathbf{H}}, y, \tilde{\cal A}, \tilde{f}_{\phi}(\bz))$ = $Distill( {\mathbf{H}}, y, {\cal A},f_{\phi}(\bz))$
        \State \( \mathcal{M} \gets reservoir(\mathcal{M}, (\tilde{\mathbf{H}}, y, \tilde{\cal A},\tilde{f}_{\phi}(\bz))) \).
    \EndFor
\EndFor
\end{algorithmic}
\end{algorithm}
\section{Experiments} \label{sec:exp}
In this section we present experiments demonstrating the efficacy of our MIL methods on CL for WSIs.

\subsection{Setup} \label{sec:exp-setup}
{\bf Datasets.}  
For class incremental learning (CIL), we utilized a skin cancer dataset \citep{Esbri2024MICILMC}, as well as three publicly available datasets for WSI analysis: Camelyon16 \citep{litjens20181399,10.1001/jama.2017.14585}, TCGA-LUNG \citep{cancer2014comprehensive}, and TCGA-RCC \citep{cancer2013comprehensive}. 
These datasets represent diverse challenges in histopathology and support the evaluation of our method's ability to generalize across different cancer types and data distributions.  Details about the datasets and preprocessing steps are in the Supplemental.

We utilized two primary groups of datasets for class-incremental learning tasks. First, we followed \citep{Esbri2024MICILMC}, which divides the skin cancer dataset into three class-incremental tasks. Each task within this group comprises two distinct class labels. The second group (denoted as Camelyon-TCGA) consists of the Camelyon16, TCGA-LUNG, and TCGA-RCC datasets, organized sequentially into three tasks based on dataset source: Camelyon16 $\rightarrow$ TCGA-LUNG $\rightarrow$ TCGA-RCC.
We employed cross-validation for Camelyon-TCGA.
This structured progression enables the model to incrementally learn from different domains, demonstrating its ability to generalize across varied types of cancer and histopathological data.


{\bf Models and CL methods.}
We employed CLAM \cite{lu2021data} and TransMIL \cite{shao2021transmil} as baseline MIL models. CLAM builds upon ABMIL by incorporating a multiclass categorization framework, making it particularly suitable for our experimental setup involving multiple cancer types. In contrast, TransMIL uses self-attention mechanisms instead of the gated-attention modules that our theoretical framework assumes.
Despite this architectural difference, we apply our AKD to TransMIL to study the generality of our AKD.

We compare performance with state-of-the-art CL methods for image classification, including experience replay (ER) \citep{riemer2018learning}, DER++ \citep{buzzega2020dark} and LwF \citep{li2016learning}, and a MIL-specific method, MICIL \citep{Esbri2024MICILMC}.
We also conduct joint training (JT) of the model using a single dataset of merged tasks, which is considered as the upper-bound performance of CL, 
and fine-tuning (FT) on each task dataset sequentially, which is the baseline performance.  

In the Skin Cancer dataset, each slide contains only a few hundred to a few thousand patches. Thus during CL on this dataset, we do not use our PMP to distill slides, and thus only apply our AKD or other CL methods.
For replay-based methods, we set the memory pool size to 42 and use the parameters recommended in their papers for other methods. 

On the Camelyon-TCGA, to ensure training stability, we use a balanced memory pool, which maintains an equal number of samples for each class in the memory pool, applying reservoir sampling within each class. Each WSI comprises over 10k patches, presenting a significant memory challenge for CL methods. To manage this, we evaluate replay-based methods with different memory settings of 5, 10, and 30 WSIs for comprehensive comparison.

{\bf Evaluation.}
To evaluate the performance of models across CL tasks, we primarily used balanced average accuracy (AACC) to measure the overall performance across all tasks. We also report backward transfer (BWT) and the intransience measure (IM) to quantify the ability to retain and adapt. See the Supplemental for more details.

\subsection{Comparison with CL methods}

\begin{table*}[t]
\centering
\small
\caption{CL performance of CLAM and TransMIL on Skin Cancer dataset. The best performances are highlighted as bold.}
\label{tab:skin-clam-transmil}
\begin{tabular}{l|ccc|ccc}
& \multicolumn{3}{c|}{\emph{CLAM}}
& \multicolumn{3}{c}{\emph{TransMIL}}\\
\hline
\multicolumn{1}{c|}{\textbf{Methods}} & \multicolumn{1}{c}{\textbf{AACC} $\uparrow$} & \multicolumn{1}{c}{\textbf{BWT}$\uparrow$}  & \multicolumn{1}{c|}{\textbf{IM}$\downarrow$} & 
\multicolumn{1}{c}{\textbf{AACC} $\uparrow$} & \multicolumn{1}{c}{\textbf{BWT}$\uparrow$}  & \multicolumn{1}{c}{\textbf{IM}$\downarrow$} 
\\ \hline
Joint training                       & 0.7311                              & -& - 
& 0.7647                                    & -                                       & -  
\\ \hline
Fine-tuning                          & 0.3167                              & -0.8701                            & -0.1942                                      
& 0.3000                                    & -0.9239                                 & -0.1771  
\\
LwF                                  & 0.3333                              & -0.8823                            & -0.2339         
  & 0.3000                                    & -0.9239                                 & -0.1923    
  \\
MICIL                                & 0.3000                              & -0.9239                            & -0.1923  
& 0.3000                                    & -0.8944                                 & -0.2087  
\\
ER                                   & 0.5388                              & -0.3869                            & -0.0942  
 & 0.6050                                    & -0.4041                                 & -0.1356  
 \\
DER++                                & 0.5343                              & -0.4931                            & -0.1604  
  & 0.5329                                    & -0.5124                                 & -0.1356   
  \\
MICIL w/ ER                          & 0.5285                              & -0.4523                            & -0.1275        
& 0.6275                                    & -0.3705                                 & -0.1356     
\\
Ours                                 & \textbf{0.5926}                              & -0.4056                            & -0.1604   
 & \textbf{0.6418}                                    & -0.3741                                 & -0.1523      
 \\ \hline
\end{tabular}
\end{table*}

First, we conduct experiments on the Skin Cancer dataset, and the results are presented in \cref{tab:skin-clam-transmil}. MICIL w/ ER is MICIL with experience replay.
All CL methods significantly outperform the fine-tuning approaches. Among these methods, our approach achieves AACC performance that most closely reaches the theoretical upper bound of joint training. Furthermore, the results for BWT and IM indicate that our method keeps the effectiveness in mitigating catastrophic forgetting while facilitating continual adaptation to new information.

The results on Camelyon-TCGA are presented in \cref{tab:c16_tcga_clam_transmil}.
Traditional CL methods face considerable challenges on this dataset, which includes over 2,000 large-scale WSIs. These methods struggle to perform effectively, particularly when applied to the TransMIL model, with some CL approaches even underperforming compared to the baseline fine-tuning method. This discrepancy arises because, despite their improved memory retention, traditional CL methods often fall short in adapting to new tasks, as reflected in their BWT and IM scores. In contrast, our method successfully addresses these limitations by integrating our AKD and PMP. This approach mitigates forgetting while also enhancing the model's ability to learn new tasks, resulting in performance gains of up to 37\% over other methods when applied to the CLAM model.

Finally, we note that despite the architectural differences in attention between CLAM and TransMIL, our experimental results indicate that TransMIL also works effectively with our proposed AKD, 
achieving strong outcomes across all evaluated metrics. This demonstrates the robustness and adaptability of the AKD method, even when applied to models with varying attention mechanisms.

\begin{table*}[]
\centering
\small
\caption{CL performance of CLAM and TransMIL on Camelyon-TCGA dataset. The best performances are highlighted as bold. The standard deviations of BWT and IM metrics can be found in the supplementary material.}
\label{tab:c16_tcga_clam_transmil}
\begin{tabular}{clc|ccc|ccc}
\multicolumn{1}{l}{} &
   &
  \multicolumn{1}{l|}{} &
  \multicolumn{3}{c|}{\textit{CLAM}} &
  \multicolumn{3}{c}{\textit{TransMIL}} \\ \hline
\multicolumn{1}{c|}{\textbf{CL Type}} &
  \multicolumn{1}{c|}{\textbf{Method}} &
  \textbf{Memory Size} &
  \textbf{AACC $\uparrow$} &
  \textbf{BWT $\uparrow$} &
  \textbf{IM $\downarrow$} &
  \textbf{AACC $\uparrow$} &
  \textbf{BWT $\uparrow$} &
  \textbf{IM $\downarrow$} \\ \hline
\multicolumn{1}{c|}{\multirow{2}{*}{\textbf{Baselines}}} &
  \multicolumn{1}{l|}{Joint training} &
  \multirow{2}{*}{-} &
  0.858$\pm$0.016 &
  - &
  - &
  0.818$\pm$0.027 &
  - &
  - \\
\multicolumn{1}{c|}{} &
  \multicolumn{1}{l|}{Fine-tuning} &
   &
  0.296$\pm$0.014 &
  -0.865 &
  -0.014 &
  0.290$\pm$0.002 &
  -0.751 &
  0.024 \\ \hline
\multicolumn{1}{c|}{\multirow{2}{*}{\textbf{Regularization}}} &
  \multicolumn{1}{l|}{LwF} &
  \multirow{2}{*}{-} &
  0.295$\pm$0.013 &
  -0.865 &
  -0.013 &
  0.296$\pm$0.010 &
  -0.747 &
  0.021 \\
\multicolumn{1}{c|}{} &
  \multicolumn{1}{l|}{MICIL} &
   &
  0.295$\pm$0.008 &
  -0.866 &
  -0.013 &
  0.294$\pm$0.011 &
  -0.760 &
  0.014 \\ \hline
\multicolumn{1}{c|}{\multirow{12}{*}{\textbf{Rehearsal}}} &
  \multicolumn{1}{l|}{ER} &
  \multirow{4}{*}{5 WSIs} &
  0.294$\pm$0.010 &
  -0.864 &
  -0.012 &
  0.296$\pm$0.008 &
  -0.748 &
  0.020 \\
\multicolumn{1}{c|}{} &
  \multicolumn{1}{l|}{DER++} &
   &
  0.301$\pm$0.010 &
  -0.857 &
  -0.013 &
  0.288$\pm$0.021 &
  -0.752 &
  0.025 \\
\multicolumn{1}{c|}{} &
  \multicolumn{1}{l|}{MICIL w/ ER} &
   &
  0.289$\pm$0.013 &
  -0.869 &
  -0.010 &
  0.290$\pm$0.009 &
  -0.759 &
  0.018 \\
\multicolumn{1}{c|}{} &
  \multicolumn{1}{l|}{Ours} &
   &
  \textbf{0.657$\pm$0.032} &
  -0.329 &
  -0.017 &
  \textbf{0.374$\pm$0.060} &
  -0.635 &
  0.018 \\ \cline{2-9} 
\multicolumn{1}{c|}{} &
  \multicolumn{1}{l|}{ER} &
  \multirow{4}{*}{10 WSIs} &
  0.352$\pm$0.063 &
  -0.785 &
  -0.017 &
  0.297$\pm$0.012 &
  -0.746 &
  0.021 \\
\multicolumn{1}{c|}{} &
  \multicolumn{1}{l|}{DER++} &
   &
  0.372$\pm$0.070 &
  -0.760 &
  -0.020 &
  0.299$\pm$0.015 &
  -0.738 &
  0.024 \\
\multicolumn{1}{c|}{} &
  \multicolumn{1}{l|}{MICIL w/ ER} &
   &
  0.298$\pm$0.006 &
  -0.863 &
  -0.015 &
  0.302$\pm$0.016 &
  -0.746 &
  0.016 \\
\multicolumn{1}{c|}{} &
  \multicolumn{1}{l|}{Ours} &
   &
  \textbf{0.729$\pm$0.041} &
  -0.217 &
  -0.059 &
  \textbf{0.394$\pm$0.085} &
  -0.595 &
  0.024 \\ \cline{2-9} 
\multicolumn{1}{c|}{} &
  \multicolumn{1}{l|}{ER} &
  \multirow{4}{*}{30 WSIs} &
  0.494$\pm$0.058 &
  -0.565 &
  -0.011 &
  0.308$\pm$0.011 &
  -0.728 &
  0.021 \\
\multicolumn{1}{c|}{} &
  \multicolumn{1}{l|}{DER++} &
   &
  0.449$\pm$0.048 &
  -0.645 &
  -0.020 &
  0.315$\pm$0.028 &
  -0.739 &
  0.006 \\
\multicolumn{1}{c|}{} &
  \multicolumn{1}{l|}{MICIL w/ ER} &
   &
  0.308$\pm$0.021 &
  -0.835 &
  -0.006 &
  0.298$\pm$0.019 &
  -0.739 &
  0.024 \\
\multicolumn{1}{c|}{} &
  \multicolumn{1}{l|}{Ours} &
   &
  \textbf{0.754$\pm$0.029} &
  -0.177 &
  -0.058 &
  \textbf{0.489$\pm$0.059} &
  -0.460 &
  0.018 \\ \hline
\end{tabular}
\end{table*}

\subsection{Visualizations}
\cref{fig:vis} visualizes the heatmap of CLAM attention during different CL sessions. The model's attention shifts when an MIL model is trained sequentially from different tasks. We visualized a sample in the Task 1 test set. 
The model should ideally focus on the tumor region. However, during fine-tuning, the model tends to reduce its attention to the tumor region and pay much more attention to the normal region. In contrast, for our method, the tumor region is nearly white in the difference heatmap, maintaining the original attention on the important tumor areas. Moreover, the attention changes brought by our method are much lower than those of the fine-tuning method.

\subsection{Ablation Study} \label{sec:exp-abl}
As introduced in \cref{sec:meth-pmp}, we explored several strategies for constructing the pseudo-bag representation. Here, we compare the performance of MaxMinRand with the following alternatives:

\begin{itemize}
    \item \textbf{Random}: randomly selects $K$ patches.
    \item \textbf{Max}: selects $K$ patches with the highest attention scores.
    \item \textbf{MaxMin}: selects $K/2$ patches with the highest attnetion scores, and $K/2$ with the lowest attention scores.
    \item \textbf{MaxRand}: selects $K/2$ patches each from the \textbf{Random} strategy and \textbf{Max} strategy.
\end{itemize}

\cref{tab:abl-distill} summarizes the results of these different strategies.
Surprisingly, both Random-based and MaxMin-based strategies achieve comparable performance, suggesting that AKD does not heavily rely on patches with the highest attention scores. Instead, AKD appears to emphasize the global attention distribution.

\begin{table}[]
\centering
\small
\caption{Ablation study of different distillation methods for AKD on CLAM and TransMIL in Camelyon-TCGA dataset.}
\label{tab:abl-distill}
\begin{tabular}{lcc}
\hline
\textbf{Method} & \textbf{CLAM AACC} & \textbf{TransMIL AACC} \\ \hline
Random          & 0.695$\pm$0.010        & 0.391$\pm$0.065            \\
MaxMin          & 0.679$\pm$0.025        & 0.391$\pm$0.054            \\
Max             & 0.595$\pm$0.121        & 0.340$\pm$0.008            \\
MaxRand         & 0.705$\pm$0.021        & 0.395$\pm$0.036            \\ \hline
MaxMinRand      & 0.729$\pm$0.041        & 0.394$\pm$0.085
\end{tabular}
\end{table}

We also investigated the contribution of each component in our method to the overall results. As shown in \cref{tab:abl-methods}, our proposed PMP method achieves a 37\% and 5\% improvement over the basic ER approach using entire WSI, without requiring additional storage. While the AKD method contributes a 40\% and 8\% improvement. The study clearly underscores the effectiveness of our method, as it enhances performance efficiently without imposing further storage demands, demonstrating a significant advancement in optimizing resource-constrained learning environments.

\begin{table}[]
\centering
\small
\caption{Ablation study on our components using CLAM on Camelyon-TCGA dataset.}
\label{tab:abl-methods}
\begin{tabular}{lcc}
\hline
\textbf{Method} & \textbf{CLAM AACC} & \textbf{TransMIL AACC} \\ \hline
ER              & 0.296$\pm$0.014        & 0.290$\pm$0.002            \\
AKD             & 0.692$\pm$0.037        & 0.373$\pm$0.045            \\
PMP             & 0.672$\pm$0.025        & 0.348$\pm$0.052            \\
PMP+AKD         & 0.729$\pm$0.041        & 0.394$\pm$0.085            \\ \hline
\end{tabular}
\end{table}
\section{Conclusion} \label{sec:conclusion}

In this paper, 
our analysis reveals that the main cause of catastrophic forgetting in attention MIL models is the degradation of bag-level feature representations due to changes in the attention module. To address this, we propose a pseudo-bag memory pool (PMP) method to enable replay-based CL for MIL and an attention knowledge distillation (AKD) technique to preserve the attention relationships and maintain representation quality. Experiments on 
WSI datasets demonstrate the effectiveness of PMP and AKD in significantly improving CL performance for traditional MIL architectures. This work establishes an important foundation for further research on CL for gigapixel images and weakly-supervised learning in real-world clinical applications.

\noindent{\bf Limitations.} The current limitation is that our methods still need to store past patch features. While storing features largely reduces privacy risks compared to raw images, it may still pose privacy concerns. In the future, we will explore more privacy-friendly data replay methods.

{
    \small
    \bibliographystyle{ieeenat_fullname}
    \bibliography{main}
}

\clearpage
\setcounter{page}{1}
\maketitlesupplementary

\section{Parameter Update in MIL}
\label{sec:para-mil}

In this section, we derive  (\ref{eq:grad-phi}) and (\ref{eq:grad-attn}).
The bag-level feature vector is: 
\begin{equation}
f_\phi = \boldsymbol{\phi}^\top \mathbf{Ha} 
= \boldsymbol{\phi}^\top \mathbf{z} = \sum_j \phi_j z_j,
\end{equation}
and in the binary classification setting, the model's loss function is
\begin{equation}
L = \log(1 + e^{-y f_\phi}).
\end{equation}
To obtain (\ref{eq:grad-phi}), 
\begin{align}
\frac{\partial L}{\partial \phi_j} &= \frac{\partial L}{\partial f_\phi} \cdot \frac{\partial f_\phi}{\partial \phi_j} \\
&= \Big(\frac{1}{1 + e^{-y f_\phi}} e^{-y f_\phi} (-y) \Big) \Big(z_j\Big) \\
&= \sigma(-y f_\phi)(-y z_j).
\end{align}
Thus, 
\begin{align}
\Big(\frac{\partial L}{\partial \phi_j}\Big)^2 = \sigma^2(-y f_\phi)y^2z_j^2.
\end{align}
For (\ref{eq:grad-attn}), we have
\begin{align}
\frac{\partial L}{\partial a_i} &= \frac{\partial L}{\partial f_\phi} \cdot \frac{\partial f_\phi}{\partial \mathbf{z}^\top} \cdot \frac{\partial \mathbf{z}}{\partial a_i} \\
&= \Big(\frac{1}{1 + e^{-y f_\phi}} e^{-y f_\phi})(-y)\Big) 
\Big( \boldsymbol{\phi}^\top \Big)
\Big( \mathbf{h}_i\Big)
\\
&= \sigma(-y f_\phi)(-y \boldsymbol{\phi}^\top \mathbf{h}_i).
\end{align}
Thus, 
\begin{align}
\Big(\frac{\partial L}{\partial a_i}\Big)^2 = 
    \sigma^2(-y f_\phi)y^2 (\boldsymbol{\phi}^\top \mathbf{h}_i)^2.
\end{align}

\section{Experiment Details}

We present the experimental details in this section.
\subsection{Dataset}
The skin cancer dataset comprises WSIs representing six distinct types of cutaneous soft tissue neoplasms: leiomyoma, leiomyosarcoma, dermatofibroma, dermatofibrosarcoma, spindle-cell melanoma, and fibroxanthoma. While the original, full-resolution WSIs are not publicly accessible, a curated subset is available, consisting of 600 vectorized and labeled pathology images. The Camelyon16 dataset consists of 399 slides labeled as either normal or tumor tissue. The TCGA-LUNG dataset provides data on two distinct types of lung cancer: LUAD with 534 slides, and LUSC with 512 slides. Finally, TCGA-RCC contains 940 renal cell carcinoma samples, divided among three subtypes: 121 from TCGA-KICH, 519 from TCGA-KIRC, and 300 from TCGA-KIRP.

\subsection{Data pre-processing}
For the skin cancer dataset, we used their repo \url{https://github.com/cvblab/MICIL} to get the data. For the Camelyon-TCGA dataset, we followed \citep{lu2021data}, using the automated segmentation pipeline to get the tissue regions and crop $256 \times 256$ patches at 20X magnification for each slide. We use the preset segmentation parameters for segmenting biopsy slides scanned at BWH for the Camelyon dataset and the parameters for TCGA slides for TCGA-LUNG and TCGA-RCC. The ResNet-50 model pre-trained with ImageNet is the fixed feature extractor that uses a global average pooling instead of the last convolutional module and converts each patch into a 1024-dimensional feature vector. Each task's dataset was split into 5 folds, with 4 folds used for training and the remaining fold reserved for testing. The training data is further randomly split into training and validation sets with a ratio of 4:1.

\subsection{Evaluation Metrics}
Denote $a_{t,j}$ as the accuracy on task $j$ after CL training session $t$.
\begin{equation}
\text{AACC} = \tfrac{1}{T} \sum\nolimits_{j=1}^{T} a_{T,j},
\end{equation}
where $a_{T,j}$ is the test accuracy on task $j$ after training on all $T$ tasks. 

{\bf BWT} measures how well the model retains knowledge from prior tasks as it learns new ones, with \( a_{T,j} \) indicating the accuracy on task \( j \) after training on the final task \( T \),
\begin{equation}
\text{BWT} = \tfrac{1}{T-1} \sum\nolimits_{j=1}^{T-1} (a_{T,j} - a_{j,j}).
\end{equation}
  
{\bf IM} quantifies a model's inability to learn a new task effectively compared to an ideal scenario,
\begin{equation}
\text{IM} = \tfrac{1}{T} \sum\nolimits_{j=1}^{T} (a_j^* - a_{j,j}),
\end{equation}
where \( a_j^* \) denotes the joint training accuarcy on task \( j \).

\subsection{Training}
For the skin cancer, we followed \citep{Esbri2024MICILMC} and used their experiment settings for TransMIL. We changed the learning rate to 1e-4 and epoch to 50 for CLAM. For Camelyon-TCGA, following \citep{lu2021data,shao2021transmil}, we use the Adam optimizer with a weight decay of 1e-5 and a learning rate of 2e-4. All the models are trained for 50 epochs with an early stop strategy. Training was conducted on an Nvidia RTX3090.

\subsection{Results}

\begin{figure*}[t]
    \centering    \includegraphics[width=0.49\linewidth]{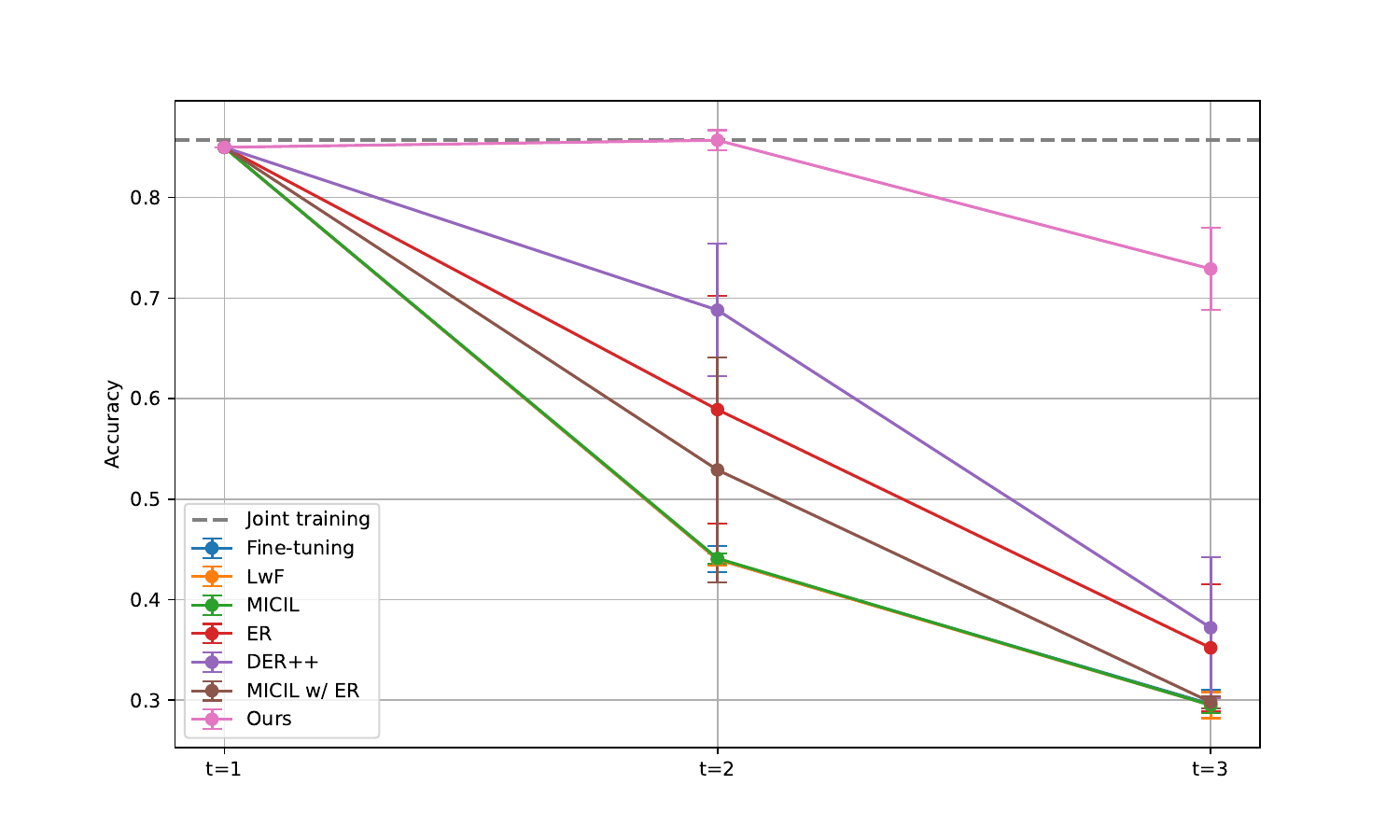}    \includegraphics[width=0.49\linewidth]{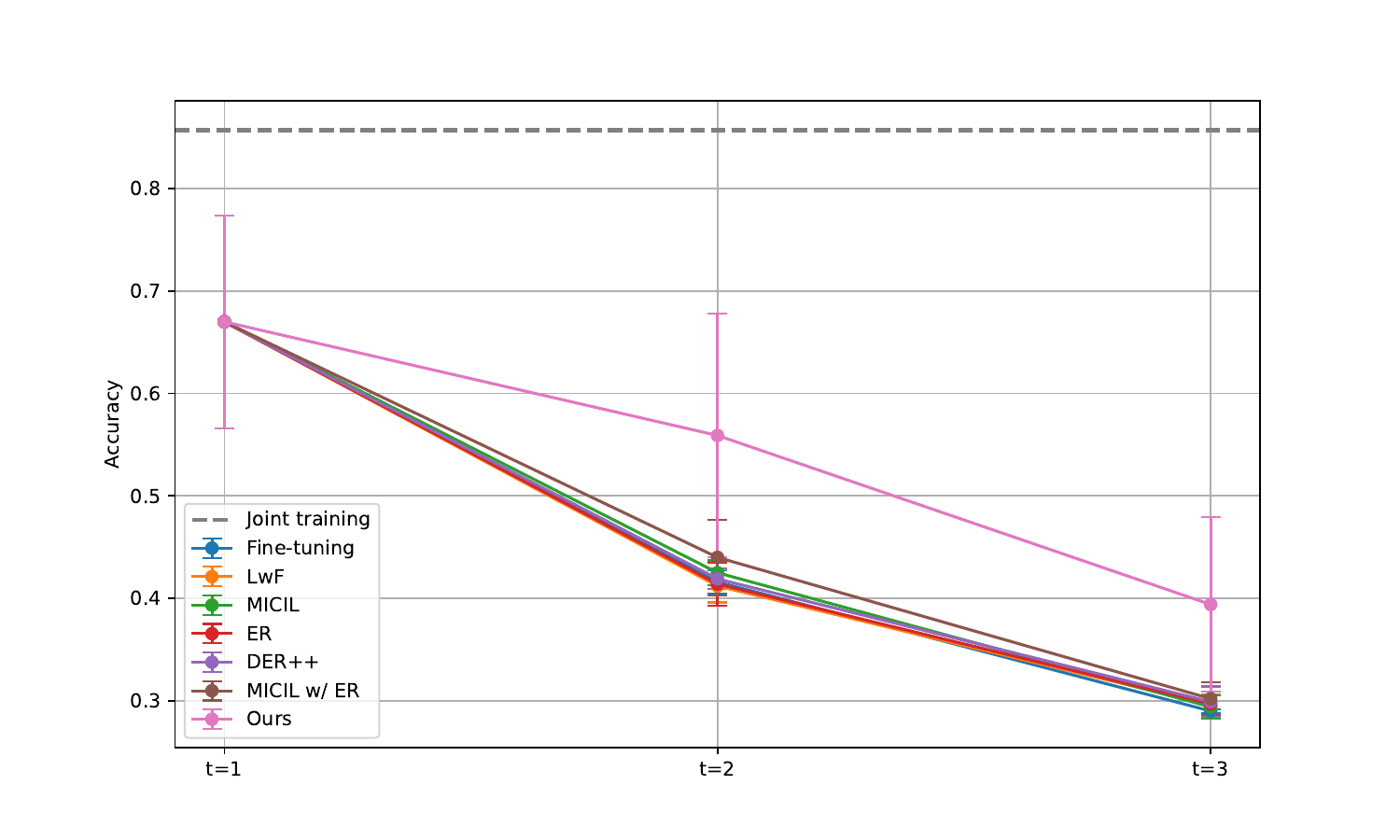}
    
    \caption{AACC performance of (left) CLAM and (right) TransMIL on Camelyon-TCGA as the number of tasks increases, i.e., the CL session $t$ increases. The gray dotted line indicates the AACC performance of joint training.}
    \label{fig:c16-tcga-line}
\end{figure*}

More detailed experimental results are shown in this section. \Cref{tab:c16_tcga_clam_transmil_full} shows the complete results including means and standard deviations. \Cref{fig:c16-tcga-line} illustrates the actual forgetting process of the model on Camelyon-TCGA with a memory setting of 10 WSIs. Our method outperforms all other methods on both $t=2$ and $t=3$ by a wide margin, exhibiting significantly less degradation.

\begin{table*}[]
\centering
\scriptsize
\caption{CL performance of CLAM and TransMIL on Camelyon-TCGA dataset. The best performances are highlighted as bold.}
\label{tab:c16_tcga_clam_transmil_full}
\begin{tabular}{clc|ccc|ccc}
\multicolumn{1}{l}{} &
   &
  \multicolumn{1}{l|}{} &
  \multicolumn{3}{c|}{\textit{CLAM}} &
  \multicolumn{3}{c}{\textit{TransMIL}} \\ \hline
\multicolumn{1}{c|}{\textbf{CL Type}} &
  \multicolumn{1}{c|}{\textbf{Method}} &
  \textbf{Memory Size} &
  \textbf{AACC $\uparrow$} &
  \textbf{BWT $\uparrow$} &
  \textbf{IM $\downarrow$} &
  \textbf{AACC $\uparrow$} &
  \textbf{BWT $\uparrow$} &
  \textbf{IM $\downarrow$} \\ \hline
\multicolumn{1}{c|}{\multirow{2}{*}{\textbf{Baselines}}} &
  \multicolumn{1}{l|}{Joint training} &
  \multirow{2}{*}{-} &
  0.858$\pm$0.016 &
  - &
  - &
  0.818$\pm$0.027 &
  - &
  - \\
\multicolumn{1}{c|}{} &
  \multicolumn{1}{l|}{Fine-tuning} &
   &
  0.296$\pm$0.014 &
  -0.865$\pm$0.013 &
  -0.014$\pm$0.019 &
  0.290$\pm$0.002 &
  -0.751$\pm$0.044 &
  0.024$\pm$0.036 \\ \hline
\multicolumn{1}{c|}{\multirow{2}{*}{\textbf{Regularization}}} &
  \multicolumn{1}{l|}{LwF} &
  \multirow{2}{*}{-} &
  0.295$\pm$0.013 &
  -0.865$\pm$0.006 &
  -0.013$\pm$0.015 &
  0.296$\pm$0.010 &
  -0.747$\pm$0.062 &
  0.021$\pm$0.045 \\
\multicolumn{1}{c|}{} &
  \multicolumn{1}{l|}{MICIL} &
   &
  0.295$\pm$0.008 &
  -0.866$\pm$0.005 &
  -0.013$\pm$0.010 &
  0.294$\pm$0.011 &
  -0.760$\pm$0.056 &
  0.014$\pm$0.031 \\ \hline
\multicolumn{1}{c|}{\multirow{12}{*}{\textbf{Rehearsal}}} &
  \multicolumn{1}{l|}{ER} &
  \multirow{4}{*}{5 WSIs} &
  0.294$\pm$0.010 &
  -0.864$\pm$0.015 &
  -0.012$\pm$0.017 &
  0.296$\pm$0.008 &
  -0.748$\pm$0.053 &
  0.020$\pm$0.030 \\
\multicolumn{1}{c|}{} &
  \multicolumn{1}{l|}{DER++} &
   &
  0.301$\pm$0.010 &
  -0.857$\pm$0.018 &
  -0.013$\pm$0.018 &
  0.288$\pm$0.021 &
  -0.752$\pm$0.050 &
  0.025$\pm$0.028 \\
\multicolumn{1}{c|}{} &
  \multicolumn{1}{l|}{MICIL w/ ER} &
   &
  0.289$\pm$0.013 &
  -0.869$\pm$0.005 &
  -0.010$\pm$0.013 &
  0.290$\pm$0.009 &
  -0.759$\pm$0.047 &
  0.018$\pm$0.036 \\
\multicolumn{1}{c|}{} &
  \multicolumn{1}{l|}{Ours} &
   &
  \textbf{0.657$\pm$0.032} &
  -0.329$\pm$0.051 &
  -0.017$\pm$0.010 &
  \textbf{0.374$\pm$0.060} &
  -0.635$\pm$0.103 &
  0.018$\pm$0.022 \\ \cline{2-9} 
\multicolumn{1}{c|}{} &
  \multicolumn{1}{l|}{ER} &
  \multirow{4}{*}{10 WSIs} &
  0.352$\pm$0.063 &
  -0.785$\pm$0.082 &
  -0.017$\pm$0.013 &
  0.297$\pm$0.012 &
  -0.746$\pm$0.062 &
  0.021$\pm$0.045 \\
\multicolumn{1}{c|}{} &
  \multicolumn{1}{l|}{DER++} &
   &
  0.372$\pm$0.070 &
  -0.760$\pm$0.097 &
  -0.020$\pm$0.021 &
  0.299$\pm$0.015 &
  -0.738$\pm$0.049 &
  0.024$\pm$0.038 \\
\multicolumn{1}{c|}{} &
  \multicolumn{1}{l|}{MICIL w/ ER} &
   &
  0.298$\pm$0.006 &
  -0.863$\pm$0.005 &
  -0.015$\pm$0.010 &
  0.302$\pm$0.016 &
  -0.746$\pm$0.049 &
  0.016$\pm$0.042 \\
\multicolumn{1}{c|}{} &
  \multicolumn{1}{l|}{Ours} &
   &
  \textbf{0.729$\pm$0.041} &
  -0.217$\pm$0.048 &
  -0.059$\pm$0.036 &
  \textbf{0.394$\pm$0.085} &
  -0.595$\pm$0.146 &
  0.024$\pm$0.030 \\ \cline{2-9} 
\multicolumn{1}{c|}{} &
  \multicolumn{1}{l|}{ER} &
  \multirow{4}{*}{30 WSIs} &
  0.494$\pm$0.058 &
  -0.565$\pm$0.081 &
  -0.011$\pm$0.016 &
  0.308$\pm$0.011 &
  -0.728$\pm$0.057 &
  0.021$\pm$0.043 \\
\multicolumn{1}{c|}{} &
  \multicolumn{1}{l|}{DER++} &
   &
  0.449$\pm$0.048 &
  -0.645$\pm$0.070 &
  -0.020$\pm$0.016 &
  0.315$\pm$0.028 &
  -0.739$\pm$0.071 &
  0.006$\pm$0.044 \\
\multicolumn{1}{c|}{} &
  \multicolumn{1}{l|}{MICIL w/ ER} &
   &
  0.308$\pm$0.021 &
  -0.835$\pm$0.015 &
  -0.006$\pm$0.009 &
  0.298$\pm$0.019 &
  -0.739$\pm$0.052 &
  0.024$\pm$0.037 \\
\multicolumn{1}{c|}{} &
  \multicolumn{1}{l|}{Ours} &
   &
  \textbf{0.754$\pm$0.029} &
  -0.177$\pm$0.041 &
  -0.058$\pm$0.035 &
  \textbf{0.489$\pm$0.059} &
  -0.460$\pm$0.059 &
  0.018$\pm$0.042 \\ \hline
\end{tabular}
\end{table*}

\begin{figure*}[t]
     \centering
     \begin{subfigure}[b]{0.43\textwidth}
         \centering         \includegraphics[width=\textwidth]{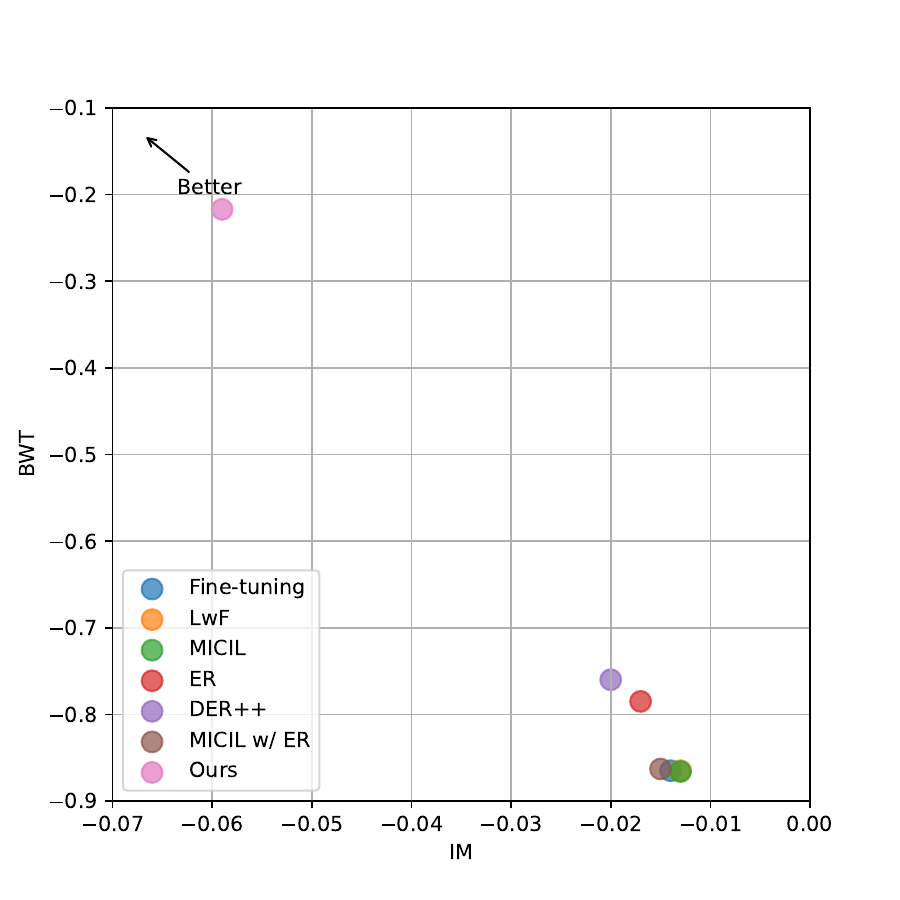}
     \end{subfigure}
     \begin{subfigure}[b]{0.43\textwidth}
         \centering
         \includegraphics[width=\textwidth]{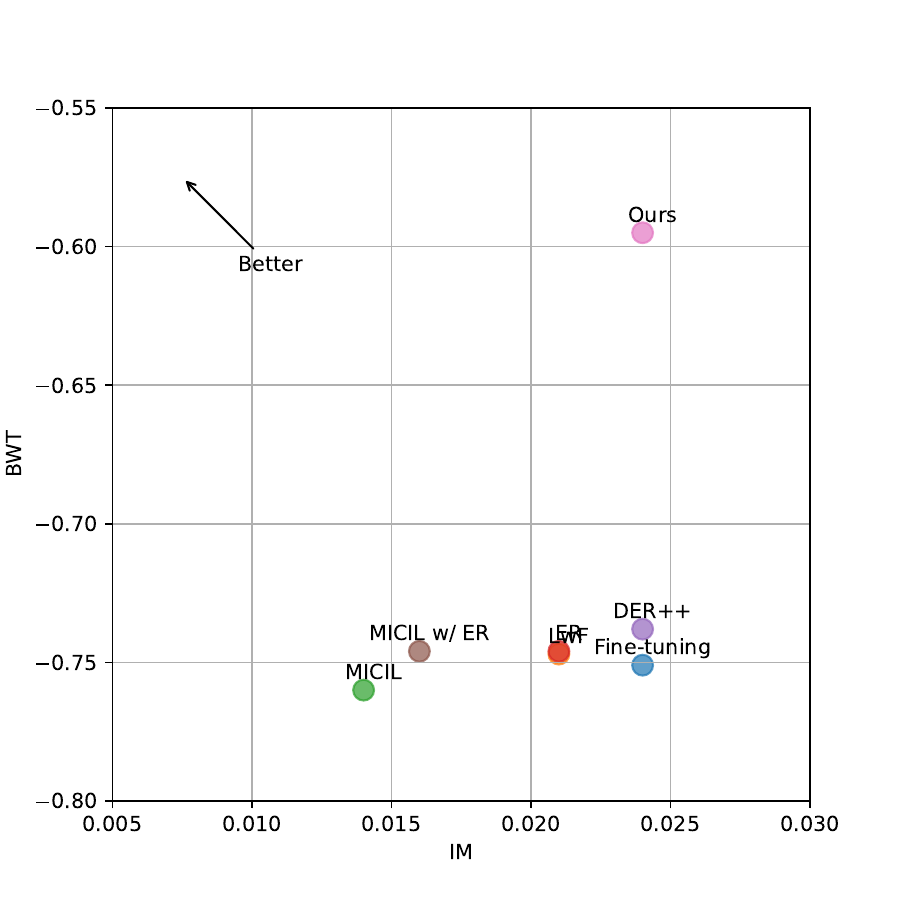}
     \end{subfigure}
        \caption{The tradeoff between  BWT and IM on Camelyon-TCGA for (left) CLAM and (right) TransMIL.
        BWT measures the amount of forgetting of previous tasks, with higher BWT indicating less forgetting.
        IM measures the ability to learn new tasks, with lower IM indicating better ability.
        Thus, better methods are closer to the upper-left corner. 
        The points represent different methods, with our method achieving the best trade-off with both models.}
        \label{fig:bwt-im}
\end{figure*}

\Cref{fig:bwt-im} provides a comparative visualization of the tradeoff between two key metrics in continual learning: BWT and IM.
BWT measures forgetting of previous tasks, with higher BWT indicating less forgetting. IM measures the model's ability to learn new tasks, with lower values indicating better ability.
For rehearsal-based methods, we only visualize results with a memory pool of 10 WSIs.
Thus, proximity to the upper-left corner indicates strong resistance to forgetting and adaptability to new tasks, i.e., models closer to this ideal area maintain knowledge from previous tasks (high BWT) while acquiring new information efficiently (low IM).
As our model is closest to the upper-left corner, our model achieves the best BWT-IM tradeoff with both MIL models.

\subsection{Compare with ConSlide}
ConSlide only released part of its code, and the WSI dataset preprocessing and partitioning steps are not released. 
Despite this, we implement the benchmark in their paper with the available information, and our results are presented in Tab.~\ref{tab:rebuttal_tcga_clam} -- our method significantly outperforms in accuracy. 

\begin{table}[ht]
\centering
\resizebox{0.9\linewidth}{!}{
\begin{tabular}{llcc}
\hline
Method & Buffer Size & ACC $\uparrow$ & BWT $\uparrow$\\
\hline
ConSlide [8] & 1100 regions $\approx$ 5 WSIs & $0.553 \pm 0.033$ & $-0.066 \pm 0.023$ \\
Ours & 5 WSIs & $0.763 \pm 0.011$ & $-0.222 \pm 0.034$ \\
\hline
ConSlide [8] & 2200 regions $\approx$ 10 WSIs & $0.594 \pm 0.053$ & $-0.092 \pm 0.026$ \\
Ours & 10 WSIs & $0.803 \pm 0.030$ & $-0.171 \pm 0.035$ \\
\hline
ConSlide [8] & 6600 regions $\approx$ 30 WSIs & $0.659 \pm 0.022$ & $-0.075 \pm 0.030$ \\
Ours & 30 WSIs & $0.868 \pm 0.028$ & $-0.071 \pm 0.026$ \\
\hline
\end{tabular}
}
\caption{CL performance of CLAM on TCGA dataset (NSCLC $\rightarrow$ BRCA $\rightarrow$ RCC $\rightarrow$ ESCA).}
\label{tab:rebuttal_tcga_clam}
\end{table}

\subsection{Visualization}

\begin{figure*}[]
    \centering
    \includegraphics[width=1\linewidth]{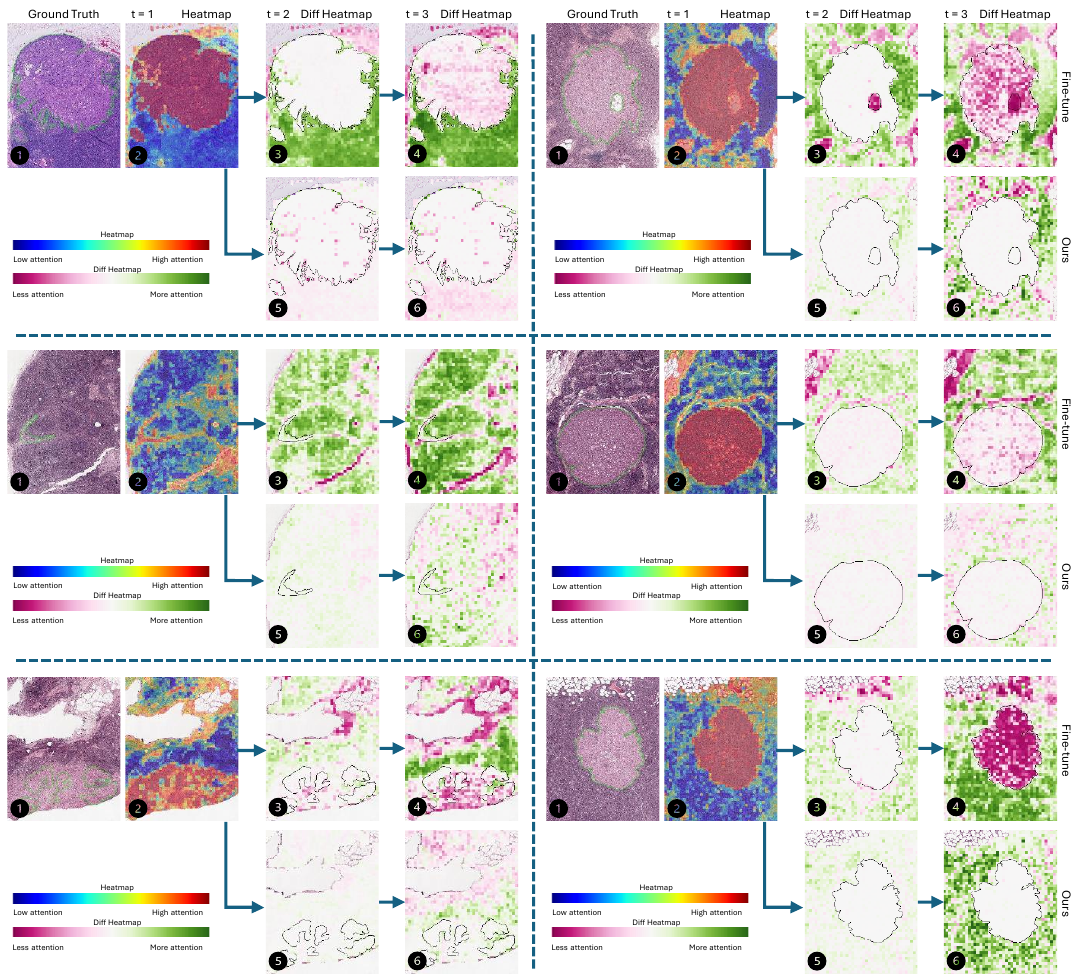}
    \caption{Additional visualizations illustrating the attention distributions across different CL sessions (\( t = 1, 2, 3 \)), using the same format as \Cref{fig:vis}.}
    \label{fig:vis-suppl}
\end{figure*}

Here we show more samples of visualization in \Cref{fig:vis-suppl}. 
The fine-tuning method exhibits limited effectiveness in maintaining the attention distribution -- attention is maintained on only a few samples in stage t=2, while in stage t=3, attention in the tumor region significantly declined across all samples. In contrast, our method consistently preserved the desired attention distribution, even at stage t=3. This demonstrates that our approach is more effective in maintaining meaningful attention to critical regions that are discriminative for classification.

\end{document}